\documentclass[11pt,a4paper]{article}
\usepackage[hyperref]{emnlp-ijcnlp-2019}
\usepackage{times}
\usepackage{latexsym}
\usepackage{graphicx}
\usepackage{bm}
\usepackage{amsmath}
\usepackage{amsfonts}
\usepackage{booktabs}
\usepackage{algpseudocode}
\usepackage{amssymb}
\usepackage{url}
\usepackage{multirow}

\graphicspath{ {./figs/} }
\def\argmax{\mathop{\rm argmax}}%

\aclfinalcopy 

\title{Improving Back-Translation \\ with Uncertainty-based Confidence Estimation}

\author{
    Shuo Wang$^\dagger$, Yang Liu$^\dagger$$^\star$\thanks{\ \ Yang Liu is the corresponding author: liuyang2011@ tsinghua.edu.cn.}, Chao Wang$^\top$, Huanbo Luan$^\dagger$, and Maosong Sun$^\dagger$\\
    $^\dagger$Institute for Artificial Intelligence\\
    State Key Laboratory of Intelligent Technology and Systems\\
    Department of Computer Science and Technology, Tsinghua University, Beijing, China\\
    Beijing National Research Center for Information Science and Technology\\
    $^\star$Beijing Advanced Innovation Center for Language Resources\\
    $^\top$6ESTATES PTE LTD, Singapore\\
}

\date{}

\begin{document}
\maketitle
\begin{abstract}
While back-translation is simple and effective in exploiting abundant monolingual corpora to improve low-resource neural machine translation (NMT), the synthetic bilingual corpora generated by NMT models trained on limited authentic bilingual data are inevitably noisy. In this work, we propose to quantify the confidence of NMT model predictions based on model uncertainty. With word- and sentence-level confidence measures based on uncertainty, it is possible for back-translation to better cope with noise in synthetic bilingual corpora. Experiments on Chinese-English and English-German translation tasks show that uncertainty-based confidence estimation significantly improves the performance of back-translation.
\footnote{The source code is available at \url{https://github.com/THUNLP-MT/UCE4BT}}
\end{abstract}

\section{Introduction}
The past several years have witnessed the rapid development of end-to-end neural machine translation (NMT) \cite{Sutskever:14,Bahdanau:15,Vaswani:17}, which leverages neural networks to map between natural languages. Capable of learning representations from data, NMT has significantly outperformed conventional statistical machine translation (SMT) \cite{Koehn:03} and been widely deployed in large-scale MT systems in the industry \cite{Wu:16,Hassan:18}.

Despite the remarkable success, NMT suffers from the data scarcity problem. For most language pairs, large-scale, high-quality, and wide-coverage bilingual corpora do not exist. Even for the top handful of resource-rich languages, the major sources of available parallel corpora are often restricted to government documents or news articles. 

\begin{figure}[!t]
    \centering
    \includegraphics[width=0.37\textwidth]{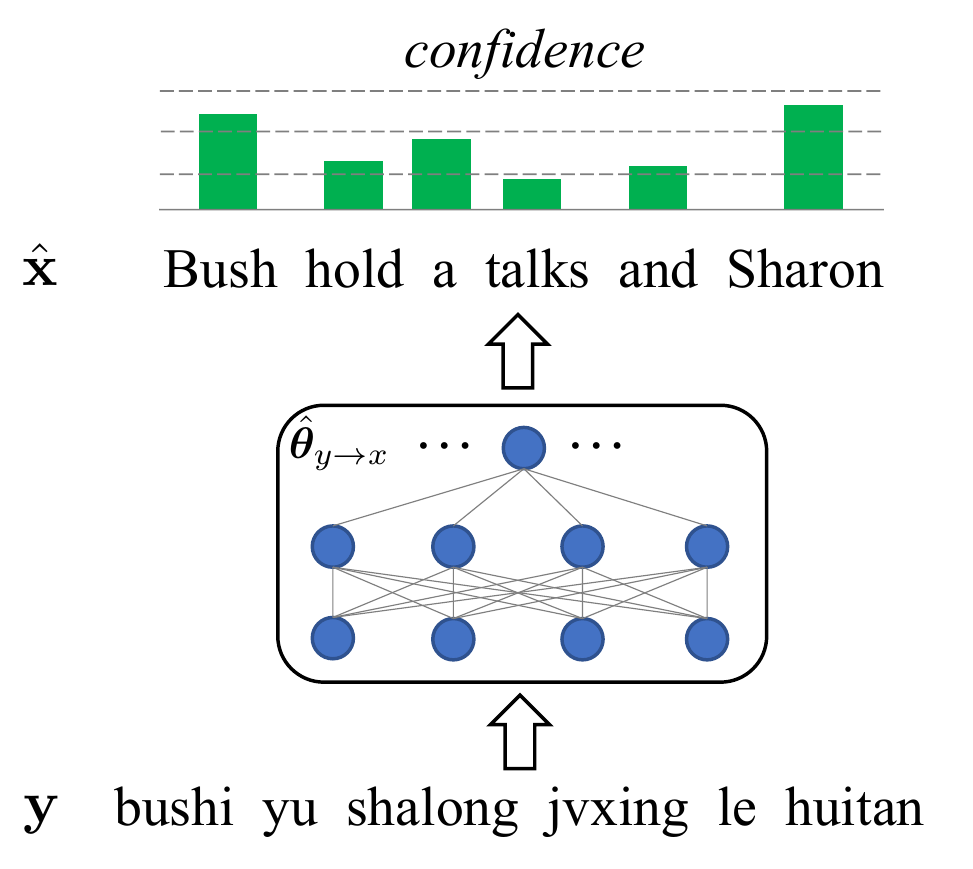}
    \caption{Confidence estimation for back-translation. Back-translation generates a source (e.g., English) sentence for a ground-truth target (e.g., Chinese) sentence. Such synthetic sentence pairs are used to train NMT models. As the model prediction (i.e., $\hat{\mathbf{x}}$) is often noisy, our work aims to quantify the prediction confidence using model uncertainty to alleviate error propagation.}
    \label{fig:example}
\end{figure}

Therefore, improving NMT under small-data training conditions has attracted extensive attention in recent years \cite{Sennrich:16,Cheng:16,Zoph:16,Chen:17,Fadaee:17,Ren:18,Lample:18}. Among them, back-translation \cite{Sennrich:16} is an important direction. Its basic idea is to use an NMT model trained on limited authentic bilingual corpora to generate synthetic bilingual corpora using  abundant monolingual data. The authentic and synthetic bilingual corpora are then combined to re-train NMT models. Due to its simplicity and effectiveness, back-translation has been widely used in low-resource language translation. However, as the synthetic corpora generated by the NMT model are inevitably noisy, translation errors can be propagated to subsequent steps and prone to hinder the performance \cite{Fadaee:18,Poncelas:2018}.

In this work, we propose a method to quantify the confidence of NMT model predictions to enable back-translation to better cope with translation errors. The central idea is to use {\em model uncertainty} \cite{Buntine:91,Gal:16,Dong:18,Xiao:19} to measure whether the model parameters can best describe the data distribution. Based on the expectation and variance of word- and sentence-level translation probabilities calculated by Monte Carlo Dropout \cite{Gal:16}, we introduce various confidence measures. 

Different from most previous quality estimation studies that require feature extraction \cite{Blatz:04,Specia2009ImprovingTC,Salehi2014ConfidenceEF} or post-edited data \cite{Kim:17,Wang:18,Ive:18} to train external confidence estimators, all our approach needs is the NMT model itself. Hence, it is easy to apply our approach to arbitrary NMT models trained for arbitrary language pairs. Experiments on Chinese-English and English-German translation tasks show that our approach significantly improves the performance of back-translation.

\section{Background}

Let $\mathbf{x} = x_1 \dots x_I$ be a source-language sentence and $\mathbf{y} = y_1 \dots y_J$ be a target-language sentence. We use $P(\mathbf{y} | \mathbf{x}, \bm{\theta}_{x \rightarrow y})$ to denote a source-to-target NMT model \cite{Sutskever:14,Bahdanau:15,Vaswani:17} parameterized by $\bm{\theta}_{x \rightarrow y}$. Similarly, the target-to-source NMT model is denoted by $P(\mathbf{x} | \mathbf{y}, \bm{\theta}_{y \rightarrow x})$.

Let $\mathcal{D}_b = \{ \langle \mathbf{x}^{(m)}, \mathbf{y}^{(m)} \rangle \}_{m=1}^{M}$ be an {\em authentic} bilingual corpus that contains $M$ sentence pairs and $\mathcal{D}_{m} = \{ \mathbf{y}^{(n)} \}_{n=1}^{N}$ be a monolingual corpus that contains $N$ target sentences. The first step of back-translation \cite{Sennrich:16} is to train a target-to-source model on the authentic bilingual corpus $\mathcal{D}_b$ using maximum likelihood estimation:

\begin{eqnarray}
\hat{\bm{\theta}}_{y \rightarrow x} = \argmax_{\bm{\theta}_{y \rightarrow x}}\Big\{ L(\mathcal{D}_b, \bm{\theta}_{y \rightarrow x}) \Big\},
\end{eqnarray}
where the log-likelihood is defined as
\begin{eqnarray}
L(\mathcal{D}_b, \bm{\theta}_{y \rightarrow x}) = \sum_{m=1}^{M} \log P(\mathbf{x}^{(m)}| \mathbf{y}^{(m)}, \bm{\theta}_{y \rightarrow x}).
\end{eqnarray}

The second step is to use the trained model $\hat{\bm{\theta}}_{y \rightarrow x}$ to translate the monolingual corpus $\mathcal{D}_m$:

\begin{eqnarray}
\hat{\mathbf{x}}^{(n)} = \argmax_{\mathbf{x}} \Big\{ P(\mathbf{x} | \mathbf{y}^{(n)}, \hat{\bm{\theta}}_{y \rightarrow x}) \Big\}, \label{eq:sent_prediction}
\end{eqnarray}
where $\hat{\mathbf{x}}^{(n)} = \hat{x}^{(n)}_1 \dots \hat{x}^{(n)}_I$. The word-level decision rule is given by
\begin{eqnarray}
\hat{x}^{(n)}_i = \argmax_{x} \Big\{ P(x | \mathbf{y}^{(n)}, \hat{\mathbf{x}}^{(n)}_{<i}, \hat{\bm{\theta}}_{y \rightarrow x}) \Big\}. \label{eq:word_prediction}
\end{eqnarray}

The resulting translations $\{ \hat{\mathbf{x}}^{(n)} \}_{n=1}^{N}$ can be combined with $\mathcal{D}_m$ to generate a {\em synthetic} bilingual corpus $\tilde{\mathcal{D}}_{b} = \{ \langle \hat{\mathbf{x}}^{(n)}, \mathbf{y}^{(n)} \rangle \}_{n=1}^{N}$. 

The third step is to train a source-to-target model $P(\mathbf{y}|\mathbf{x}, \bm{\theta}_{x \rightarrow y})$ on the combination of authentic and synthetic bilingual corpora:

\begin{eqnarray}
\hat{\bm{\theta}}_{x \rightarrow y} = \argmax_{\bm{\theta}_{x \rightarrow y}} \Big\{ L(\mathcal{D}_b \cup \tilde{\mathcal{D}}_b, \bm{\theta}_{x \rightarrow y}) \Big\} \label{eq:third_step}.
\end{eqnarray}

This three-step process can iterate until convergence \cite{Hoang:18,Cotterell2018ExplainingAG}.

A problem with back-translation is that model predictions are inevitably erroneous.
Translation errors can be propagated to subsequent steps and impair the performance of back-translation, especially when $\tilde{\mathcal{D}}_{b}$ is much larger than $\mathcal{D}_{b}$ \cite{Pinnis:17,Fadaee:18,Poncelas:2018}. Therefore, it is crucial to develop principled solutions to enable back-translation to better deal with the error propagation problem. 

\begin{figure*}[!t]
    \centering
    \includegraphics[width=1.0\textwidth]{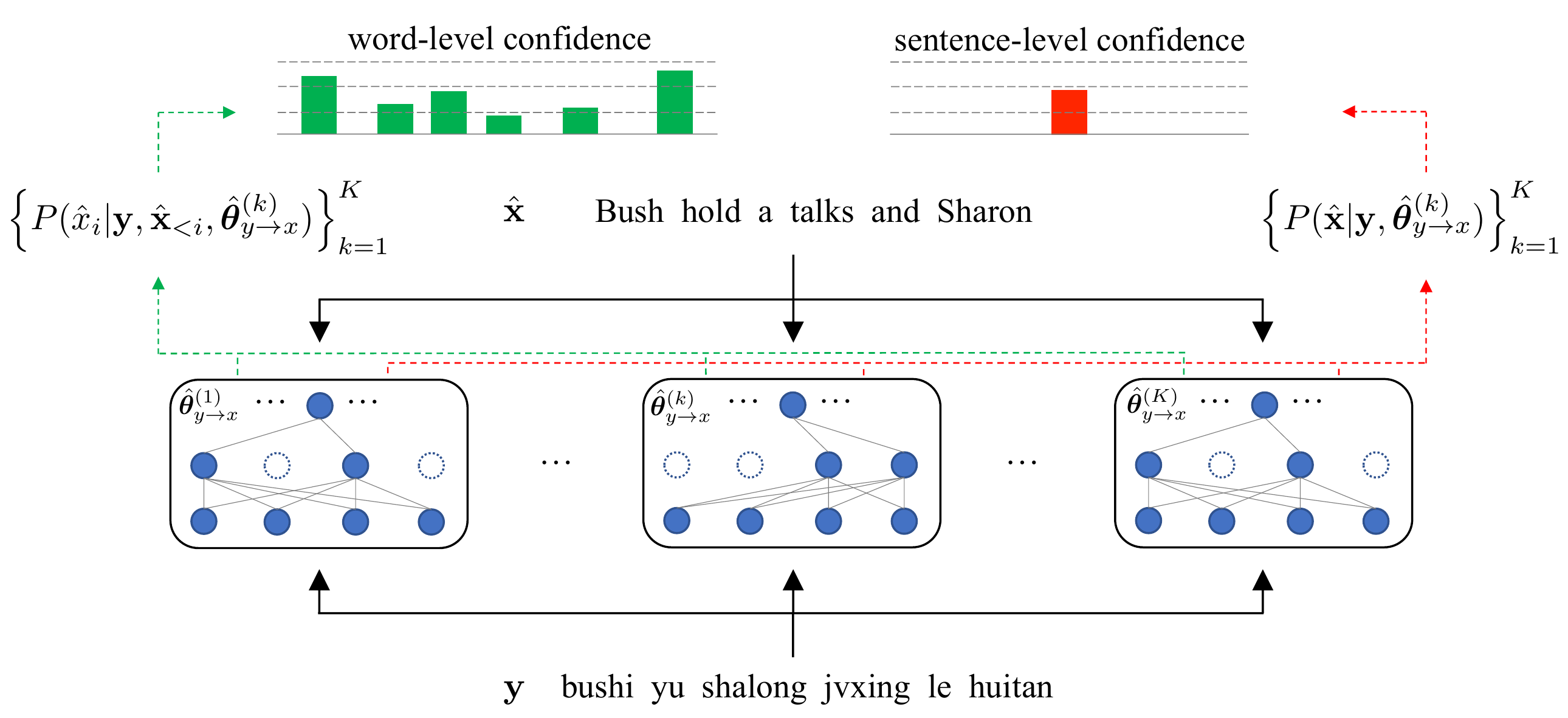}
    \caption{Illustration of uncertainty calculation. Given a target sentence $\mathbf{y}$ and the model prediction $\hat{\mathbf{x}}$, our approach treats word- and sentence-level translation probabilities as random variables and uses Monte Carlo Dropout to draw samples. These samples are used to calculate the expectations and variances of translation probabilities.}
    \label{fig:dropout}
\end{figure*}

\section{Approach}

This work aims to 
find solutions to the two following problems:
\begin{enumerate}
\item How to quantify the confidence of model predictions at both word and sentence levels?
\item How to leverage confidence to improve back-translation?
\end{enumerate}
Section \ref{sec:uncertainty} introduces how to calculate model uncertainty, which lays a foundation for designing uncertainty-based word- and sentence-level confidence measures in Section \ref{sec:measures}. Section \ref{sec:training} describes confidence-aware training for NMT models on noisy bilingual corpora.

\subsection{Calculating Uncertainty} \label{sec:uncertainty}
Uncertainty quantification, which quantifies how confident a certain mapping is with respect to different inputs, has made significant progress due to the recent advances in Bayesian deep learning \cite{Kendall:15,Gal:16,Kendall:17,Xiao:19,oh:18modeling,geifman:18biasreduced,lee:18bayesian}. 
In this work, we aim to calculate {\em model uncertainty} \cite{Kendall:17,Dong:18,Xiao:19}, which measures whether a model can best describe the data distribution, for NMT using approximate inference methods widely used in Bayesian neural networks.

Given the authentic bilingual corpus $\mathcal{D}_b$, Bayesian neural networks aim at finding the posterior distribution over model parameters $P(\bm{\theta}_{y \rightarrow x} | \mathcal{D}_b)$. 
With a target sentence $\mathbf{y}$ in the monolingual corpus $\mathcal{D}_m$ and its translation $\hat{\mathbf{x}}$, the translation probability is given by
\begin{eqnarray}
P(\hat{\mathbf{x}} | \mathbf{y}, \mathcal{D}_b) \quad \quad \quad \quad \quad \quad \quad \quad \quad \quad \nonumber \\
= \int P(\hat{\mathbf{x}} | \mathbf{y}, \bm{\theta}_{y \rightarrow x}) P(\bm{\theta}_{y \rightarrow x} | \mathcal{D}_b) d \bm{\theta}_{y \rightarrow x}.
\end{eqnarray}

In particular, we are interested in calculating the variance of the distribution $P(\hat{\mathbf{x}} | \mathbf{y}, \bm{\theta}_{y \rightarrow x})$ that reflects our ignorance over model parameters, which is referred to as {\em model uncertainty}.
As exact inference is intractable, a number of variational inference methods \cite{Graves:11PracticalVI,Blundell:15,Gal:16} have been proposed to find an approximation to $P(\bm{\theta}_{y \rightarrow x} | \mathcal{D}_b)$. In this work, we leverage the widely used Monte Carlo Dropout \cite{Gal:16} to obtain samples of word- and sentence-level translation probabilities. 

Figure \ref{fig:dropout} illustrates the key idea of our approach. Given an authentic target sentence $\mathbf{y}$, an NMT model made its prediction $\hat{\mathbf{x}}$ via a standard decoding process (see Eq. (\ref{eq:sent_prediction}) and Eq. (\ref{eq:word_prediction})).
To quantify how confident the model was when making the prediction, our approach treats word- and sentence-level translation probabilities as random variables.
\footnote{Unlike prior studies that calculate model uncertainty during inference \cite{Xiao:19}, our approach computes uncertainty after the NMT model has made the prediction for two reasons. First, our goal is to quantify the confidence of model prediction rather than using uncertainty to improve  model prediction. Second, using Monte Carlo Dropout during decoding is very slow because of the autoregressive property of standard NMT models.}
Drawing samples can be done by randomly deactivating part of neurons of the NMT model and re-calculating translation probabilities while keeping $\mathbf{y}$ and $\hat{\mathbf{x}}$ fixed.
This stochastic feedforward is repeated $K$ times and generates $K$ samples
for both word- and sentence-level translation probabilities, respectively. We use $\hat{\bm{\theta}}^{(k)}_{y \rightarrow x}$ to denote the model parameters derived from $\hat{\bm{\theta}}_{y \rightarrow x}$ by deactivation in the $k$-th pass. 

Intuitively, if the variance of translation probability is low, it is highly likely that the model was confident in making the prediction.
Given $K$ samples $\{P(\hat{\mathbf{x}}|\mathbf{y}, \hat{\bm{\theta}}^{(k)}_{y \rightarrow x})\}_{k=1}^{K}$, the expectation of sentence-level translation probability can be approximated by

\begin{eqnarray}
\mathbb{E} \Big[ P(\hat{\mathbf{x}} | \mathbf{y}, \hat{\bm{\theta}}_{y \rightarrow x}) \Big] \approx \frac{1}{K} \sum_{k=1}^{K} P(\hat{\mathbf{x}} | \mathbf{y}, \hat{\bm{\theta}}^{(k)}_{y \rightarrow x}). \label{eq:expectation}
\end{eqnarray}

The variance of sentence-level translation probability can be approximated by
\begin{eqnarray}
\mathrm{Var}\Big[ P(\hat{\mathbf{x}} | \mathbf{y}, \hat{\bm{\theta}}_{y \rightarrow x}) \Big] \quad \quad \quad \quad \quad \quad \quad \quad \quad \quad \nonumber \\
\approx \frac{1}{K} \sum_{k=1}^{K} P(\hat{\mathbf{x}} | \mathbf{y}, \hat{\bm{\theta}}^{(k)}_{y \rightarrow x})^2  - \mathbb{E} \Big[ P(\hat{\mathbf{x}} | \mathbf{y}, \hat{\bm{\theta}}_{y \rightarrow x}) \Big]^2, \label{eq:variance}
\end{eqnarray}
which is also referred to as {\em model uncertainty}. 

The expectation and variance of word-level translation probabilities can also be calculated similarly using $K$ samples.



\subsection{Confidence Measures} \label{sec:measures}

We use $C(\mathbf{y}, \hat{\mathbf{x}}_{<i}, \hat{x}_i, \hat{\bm{\theta}}_{y \rightarrow x})$ to denote the {\em word-level confidence} for the model to generate $\hat{x}_i$ and $C(\mathbf{y}, \hat{\mathbf{x}}, \hat{\bm{\theta}}_{y \rightarrow x})$ to the denote the {\em sentence-level confidence} for the model to generate $\hat{\mathbf{x}}$.

Intuitively, when making predictions, the more confident an NMT model is, the higher expectation and lower variance of translation probability are. For comparison reasons, we used the following four types of confidence measures at the sentence level in our experiments: 

\begin{enumerate}
\item {\em Predicted translation probability} (\textproc{PTP}). The translation probability of model prediction during standard decoding (Eq. (\ref{eq:sent_prediction})):
\begin{eqnarray}
C_{\mathrm{PTP}}(\mathbf{y}, \hat{\mathbf{x}}, \hat{\bm{\theta}}_{y \rightarrow x}) = P(\hat{\mathbf{x}} | \mathbf{y}, \hat{\bm{\theta}}_{y \rightarrow x}). \quad \quad 
\end{eqnarray}

\item {\em Expected translation probability} (\textproc{EXP}). The expectation of translation probability:
\begin{eqnarray}
C_{\mathrm{EXP}}(\mathbf{y}, \hat{\mathbf{x}}, \hat{\bm{\theta}}_{y \rightarrow x}) = \mathbb{E} \Big[ P(\hat{\mathbf{x}} | \mathbf{y}, \hat{\bm{\theta}}_{y \rightarrow x}) \Big].
\end{eqnarray}

\item {\em Variance of translation probability} (\textproc{VAR}). The variance of translation probability:
\begin{eqnarray}
\label{eq:var}
C_{\mathrm{VAR}}(\mathbf{y}, \hat{\mathbf{x}}, \hat{\bm{\theta}}_{y \rightarrow x}) \quad \quad \quad \quad \  \nonumber \\
= \Big( 1 - \mathrm{Var}\Big[ P(\hat{\mathbf{x}} | \mathbf{y}, \hat{\bm{\theta}}_{y \rightarrow x}) \Big] \Big)^\alpha.
\end{eqnarray}

\item {\em Combination of expectation and variance} (\textproc{CEV}). The combination of expectation and variance:
\begin{eqnarray}
\label{eq:cve}
C_{\mathrm{CEV}}(\mathbf{y}, \hat{\mathbf{x}}, \hat{\bm{\theta}}_{y \rightarrow x}) \quad \quad \quad \quad \ \ \ \nonumber \\
= \Bigg( 1 - \frac{\mathrm{Var}\big[ P(\hat{\mathbf{x}} | \mathbf{y}, \hat{\bm{\theta}}_{y \rightarrow x}) \big]}{\mathbb{E} \big[ P(\hat{\mathbf{x}} | \mathbf{y}, \hat{\bm{\theta}}_{y \rightarrow x}) \big]} \Bigg)^\beta.
\end{eqnarray}
\end{enumerate}
where $\alpha$ and $\beta$ are hyper-parameters to control the gap between confidence values of sentences of different quality. Larger values of $\alpha$ and $\beta$ lead to bigger gaps. \footnote{Note that all confidence measures are between 0 and 1. Clearly, both the expectation and variance of a probability are between 0 and 1. It can be proved that the variance of a probability is no greater than the corresponding expectation. As a result, $C_{\mathrm{CEV}}(\cdot)$ is also between 0 and 1.}

In Eq. (\ref{eq:cve}), our approach tries to combine the merits of expectation and variance by using variance divided by expectation because smaller variance and bigger expectation are expected to result in higher confidence. 
There may exist more sophisticated ways to estimate prediction confidence using model uncertainty \cite{Dong:18}. As we find that the measures mentioned above are easy-to-implement and prove to be effective in our experiments, we leave the investigation of more complex confidence measures for future work.




The word-level confidence measures can be defined similarly.

\begin{figure*}[!t]
    \centering
    \includegraphics[width=1.0\textwidth]{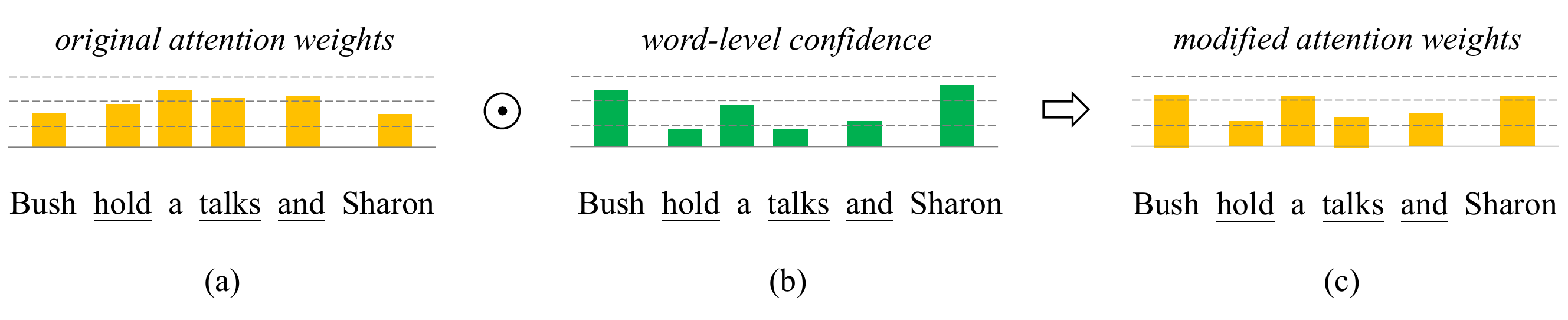}
    \caption{Using word-level confidence in confidence-aware training. The basic idea is to use confidence to modify attention weights to pay less attention to erroneous words highlighted in underline. (a) The original attention weights of the NMT model; (b) the word-level confidence of the noisy source sentence; (c) the attention weights modified by the word-level confidence, which  focus more on words with high confidence. 
    $\odot$ is a broadcast product.
    See Eq. (\ref{eq:word-levelCAT}) for details.}
    \label{fig:wordconf}
\end{figure*}

\subsection{Confidence-aware Training for NMT} \label{sec:training}

We propose confidence-aware training for NMT to enable NMT to make better use of noisy data.
Word- and sentence-level confidence measures are complementary: while word-level confidence can provide more fine-grained information than the sentence-level counterpart, it is unable to cope with word omission errors that can only be captured at the sentence level. As a result, our approach incorporates both word- and sentence-level confidence measures into the training process.
\footnote{Instead of applying confidence estimation to the second pass of decoding \cite{Luong2017FindTE}, we directly integrate confidence scores into the training process. These two kinds of methods are complementary.}

\subsubsection*{Using Sentence-level Confidence}

It is easy to integrate sentence-level confidence into back-translation by modifying the likelihood function in Eq. (\ref{eq:third_step}):

\begin{eqnarray}
&& L(\mathcal{D}_b \cup \tilde{\mathcal{D}}_b, \bm{\theta}_{x \rightarrow y}) \nonumber \\
&=& \sum_{m=1}^{M} \log P(\mathbf{y}^{(m)}|\mathbf{x}^{(m)}, \bm{\theta}_{x \rightarrow y}) + \nonumber \\
&& \sum_{n=1}^{N} C(\mathbf{y}^{(n)}, \hat{\mathbf{x}}^{(n)}, \hat{\bm{\theta}}_{y \rightarrow x}) \times \nonumber \\
&& \quad \ \ \ \log P(\mathbf{y}^{(n)}|\hat{\mathbf{x}}^{(n)}, \bm{\theta}_{x \rightarrow y}).
\end{eqnarray} 

Serving as a weight assigned to each synthetic sentence pair, sentence-level confidence is expected to help to minimize the negative effect of estimating parameters on sentences with lower confidence. Note that the confidence of an authentic sentence pair in $\mathcal{D}_b$ is 1.

\subsubsection*{Using Word-level Confidence}

As the source side instead of the target side of the synthetic bilingual corpus is noisy, word-level confidence cannot be integrated into back-translation in a similar way to sentence-level confidence. This is because the word-level confidence associated with each source word does not get involved in backpropagation during training.

Alternatively, we build a real-valued word-level confidence vector:
\begin{eqnarray}
\mathbf{c} = \Big\{ C\big(\mathbf{y}^{(n)}, \hat{\mathbf{x}}^{(n)}_{<i}, \hat{x}_i, \hat{\bm{\theta}}_{y \rightarrow x}\big) \Big\}_{i=1}^{I}.
\end{eqnarray}

Due to the wide use of attention \cite{Bahdanau:15,Vaswani:17} in NMT, we use the confidence vector $\mathbf{c} \in \mathbb{R}^{1 \times I}$ to modify attention weights and enable the model to focus more on words with high confidence. Figure \ref{fig:wordconf} shows an example. Figure \ref{fig:wordconf}(a) gives a source sentence in the synthetic bilingual corpus, in which erroneous words ``hold'', ``talks'', and ``and'' receive high attention weights, deteriorating the parameter estimation on this sentence pair. By multiplying with word-level confidence (Figure \ref{fig:wordconf}(b)), the weights are modified to pay less attention to erroneous words (Figure \ref{fig:wordconf}(c)).

More formally, the modified attention function is given by
\begin{eqnarray} \label{eq:word-levelCAT}
&& \mathrm{Attention}(\mathbf{Q}, \mathbf{K}, \mathbf{V}, \mathbf{c})  \nonumber \\
&=& \left ( \mathrm{softmax} \Big( \frac{\mathbf{Q} \mathbf{K}^{\top}}{\sqrt{D}} \Big) \odot \mathbf{c} \right) \mathbf{V},
\end{eqnarray}
where $\mathbf{Q} \in \mathbb{R}^{I \times D}$, $\mathbf{K} \in \mathbb{R}^{I \times D}$, and $\mathbf{V} \in \mathbb{R}^{I \times D}$ are query, key, and value matrices and $D$ is the hidden size. $\odot$ is a broadcast product. 

Since the integration of sentence- and word-level confidence measures are independent of each other, it is easy to use both of them in back-translation.

\section{Experiments}

\subsection{Setup}

We evaluated our approach on Chinese-English and English-German translation tasks. The evaluation metric is BLEU \cite{Papineni2001BleuAM} as calculated by the \verb|multi-bleu.perl| script. We use the paired bootstrap resampling \cite{Koehn2004Statistical} for significance testing.

For the Chinese-English task, the training set contains 1.25M sentence pairs from LDC\footnote{The training set includes LDC2002E18, LDC2003E07, LDC2003E14, part of LDC2004T07, LDC2004T08 and LDC2005T06.} with 27.8M Chinese words and 34.5M English words.
To build the monolingual corpus for back-translation, we extracted the English side of the training set of the WMT 2017 Chinese-English news translation task.
After removing sentences longer than 256 words, we randomly selected 10M English sentences as the monolingual corpus. NIST06 is used as the development set and NIST02, 03, 04, 05, and 08 datasets as test sets. 

For the English-German task, we used the dataset of the WMT 2014 English-German translation task. The training set consists of 4.47M sentence pairs with 116M English words and 110M German words. 
We randomly selected 4.5M German sentences from the 2012 News Crawl corpus of WMT 2014  to construct the monolingual corpus 
for back-translation.
We use newstest 2013 as the development set and newstest 2012, 2014, and 2015 as test sets.

\begin{table}[!t]
\begin{center}
\begin{tabular}{c|cc}
\toprule
Measure & BLEU & $\Delta$  \\ 
\midrule
- & 46.23 & - \\
\midrule
PTP & 45.41 & -0.82  \\
EXP & 45.22 & -1.01 \\
VAR & 46.77 & +0.54 \\
CEV & \textbf{47.05} & \textbf{+0.82} \\
\bottomrule
\end{tabular}
\end{center}
\caption{Comparison of confidence measures.} \label{tab:type}
\end{table}

Chinese sentences were segmented by an open-source toolkit \verb|THULAC|\footnote{\url{https://github.com/thunlp/THULAC-Python}}
. German and English sentences were tokenized by the tokenizer in \verb|Moses| \cite{Koehn2007MosesOS}. 
We used byte pair encoding \cite{Sennrich2016NeuralMT} to perform sub-word segmentation with 32k merge operations for Chinese-English and 35k merge operations for English-German. 
Sentence pairs are batched together by approximate length and each batch has roughly 25,000 source and target tokens.
We distinguish between three kinds of translations of the monolingual corpus:
\begin{enumerate}
\item \textproc{None}: there is no translation and only the authentic bilingual corpus is used;
\item \textproc{Search}: the translations are generated by beam search \cite{Sennrich:16};
\item \textproc{Sample}: the translations are generated by sampling \cite{Edunov:18}.
\end{enumerate}

As neural quality estimation \cite{Kim:17,Wang:18} can also give word- and sentence-level confidences for the output of NMT models when labeled data is available, we distinguish between two kinds of confidence estimation methods:
\begin{enumerate}
\item \textproc{NeuralQe}: the confidences are given by an external neural quality estimator;
\item \textproc{Uncertainty}: the proposed uncertainty-based confidence estimation method.
\end{enumerate}

For \textproc{NeuralQe}, we used the Predictor-Estimator architecture \cite{Kim:17} implemented by \verb|OpenKiwi|
\cite{openkiwi}, which is an open source software officially recommended by the QE shared task of WMT. Following the guide of \verb|OpenKiwi|, we used a German-English parallel corpus containing 2.09M sentence pairs to train the predictor and a post-edited corpus containing 25k sentence triples to train the estimator. All the data used to train QE models are provided by WMT.
As there are no post-edited corpora for the Chinese-English task, \textproc{NeuralQe} can only be used in the English-German task in our experiments. For \textproc{NeuralQe}, both word- and sentence- level quality scores were considered.

We implemented our method on the top of \verb|THUMT|
\cite{THUMT}.
The NMT model we use is Transformer \cite{Vaswani:17}. 
We used the base model for the Chinese-English task and the big model for the English-German task.  
We used the Adam optimizer \cite{Kingma2015AdamAM} with $\beta_{1}=0.9$, $\beta_{2}=0.98$ and $\epsilon=10^{-9}$ to optimize model parameters.
 We used the same warm-up strategy for learning rate as \newcite{Vaswani:17} with $\rm warmup\_steps=4,000$.
During training, the hyper-parameter of label smoothing was set as $\epsilon_{ls}=0.1$ \cite{Szegedy:2016,Pereyra:2017}. 
During training and the Monte Carlo Dropout process, the hyper-parameter of dropout was set to 0.1 and 0.3 for Transformer base and big models, respectively. $K$ was set to 20.
Through experiments, we find our method works best when the $\alpha$ and $\beta$ are set to 2.
All experiments were conducted on 8 NVIDIA GTX 1080Ti GPUs.

\begin{table}[!t]
\begin{center}
\begin{tabular}{cc|cc}
\toprule
Word & Sentence & BLEU & $\Delta$\\ 
\midrule
$\times$ & $\times$ & 46.23 & - \\
\midrule
$\times$ & $\surd$ & 46.42 & +0.19 \\
$\surd$ & $\times$ & 46.98 & +0.75 \\
$\surd$ & $\surd$ & \textbf{47.05} & \textbf{+0.82}\\
\bottomrule
\end{tabular}
\end{center}
\caption{Comparison between word- and sentence-level CEV confidence measures.} \label{tab:level} 
\end{table}

\begin{table*}[!ht]
\begin{center}
\begin{tabular}{l|c|l|lllll|l}
\toprule
Data & CE
& MT06 & MT02 & 
MT03 & MT04 &
MT05 & MT08 & All \\
\midrule 
\textproc{None} & - &
45.05 &
45.09 &
44.79 &
46.07 &
44.34 &
35.52 &
43.50 \\
\midrule
\multirow{2}{*}{\textproc{Search}} & - &
46.23 &
45.85 &
45.37 &
46.77 &
46.28 &
37.69 &
44.76 \\
 & U &
\textbf{47.05}$^{++}$ &
\textbf{48.06}$^{++}$ &
\textbf{46.44}$^{++}$ &
\textbf{47.59}$^{++}$ &
\textbf{47.03}$^{++}$ &
\textbf{38.02}$^{+}$ &
\textbf{45.72}$^{++}$ \\
\midrule
\multirow{2}{*}{\textproc{Sample}} & - &
46.69 &
46.98 &
45.62 &
46.97 &
46.29 &
37.28 &
44.96 \\
 & U &
\textbf{46.78} &
46.75 &
\textbf{46.53}$^{\ddag\ddag}$ &
\textbf{47.70}$^{\ddag\ddag}$ &
\textbf{47.48}$^{\ddag\ddag}$ &
36.99 &
\textbf{45.37}$^{\ddag\ddag}$ \\
\bottomrule
\end{tabular}
\end{center}
\caption{BLEU scores on the NIST Chinese-English translation task.
The ratio of authentic data to synthetic data is 1:1.
\textproc{None}: only the authentic bilingual corpus is used.
\textproc{Search}: the translations of the monolingual corpus are generated by beam search \cite{Sennrich:16}.
\textproc{Sample}: the translations of the monolingual corpus are generated by sampling \cite{Edunov:18}.
``CE'': confidence estimation method.
``U'': the proposed uncertainty-based confidence estimation.
``All'': the combination of all test sets. 
``$+$'': significantly better than \textproc{Search} without CE ($p < 0.05$). ``$++$'': significantly better than \textproc{Search} without CE ($p < 0.01$). ``$\ddag\ddag$'': significantly better than \textproc{Sample} without CE ($p < 0.01$).
} \label{tab: main zhen}
\end{table*}

\subsection{Comparison of Confidence Measures}

Table \ref{tab:type} shows the comparison of confidence measures on the Chinese-English development set. We find that using either the translation probabilities outputted by the model (i.e., ``PTP'') or the expectation of translation probabilities (i.e., ``EXP'') deteriorates the translation quality, which suggests that translation probabilities themselves can not help NMT models better cope with synthetic data. In contrast, using the variance or model uncertainty (i.e., ``VAR'') increases translation quality. Combining variance and expectation (i.e., ``CEV'') leads to a further improvement. 
In the following experiments, we use CEV as the default setting.

\subsection{Comparison between Word- and Sentence-level Confidence Measures}

Table \ref{tab:level} shows the comparison between word- and sentence-level CEV (i.e., combination of expectation and variance) confidence measures on the Chinese-English development set. It is clear that using either sentence-level or word-level confidence measures improves the translation performance. Thanks to more fine-grained quantification of uncertainty, using word-level confidence achieves a higher BLEU score than using sentence-level confidence. 
Combining the sentence- and word-level of confidences leads to a further improvement, suggesting that they are complementary to each other. In the following experiments, we use the combination of word- and sentence-level confidences as the default setting.

\subsection{Main Results}\label{sec:mainres}

\subsubsection*{The Chinese-English Task}

Table \ref{tab: main zhen} shows the results of the Chinese-English task. Back-translation, either generating translations using beam search (i.e., \textproc{Search}) or using sampling (i.e., \textproc{Sample}), does lead to significant improvements over using only the authentic bilingual corpus (i.e., \textproc{None}). We find that \textproc{Sample} is more effective than \textproc{Search}, which confirms the finding of \citet{Edunov:18}. Using uncertainty-based confidence (i.e., ``U'') significantly improves over both \textproc{Search} and \textproc{Sample} on the combination of all test sets ($p<0.01$). As there is no Chinese-English labeled data to train neural quality estimation models, we did not report the result of \textproc{NeuralQe} in this experiment.



\subsubsection*{The English-German Task}

Table \ref{tab: main ende} shows the results of the English-German task. We find that using quality estimation, either \textproc{NeuralQe} (i.e., ``N'') or \textproc{Uncertainty} (i.e., ``U''), improves over \textproc{Search} and \textproc{Sample}.  \textproc{Uncertainty} even achieves better performance than \textproc{NeuralQe}, although \textproc{NeuralQe} uses additional labeled training data. As \textproc{NeuralQe} heavily relies on post-edited corpora and labeled data to train QE models, it can only be used in a handful of language pairs. In contrast, it is easier to apply our approach to arbitrary language pairs since it does not need any labeled data to estimate confidence.

\begin{table*}[!ht]
\begin{center}
\begin{tabular}{l|c|l|lll|l}
\toprule
Data & CE &
news2013 & news2012 & 
news2014 & news2015 & All\\
\midrule
\textproc{None} & - &
26.57 &
22.09 &
28.42 &
30.26 &
26.31 \\
\midrule
\multirow{3}{*}{\textproc{Search}} & - &
27.09 &
23.10 &
29.45 &
30.10 &
27.04 \\
& N &
27.58 &
23.91 &
30.61 &
31.87 &
28.18 \\
& U &
\textbf{27.89}$^{++*}$ &
23.75$^{++}$ &
\textbf{31.00}$^{++*}$ &
\textbf{31.98}$^{++}$ &
\textbf{28.28}$^{++}$ \\
\midrule
\multirow{3}{*}{\textproc{Sample}} & 
- &
27.30 &
23.37 &
30.11 &
31.51 &
27.70 \\
& N &
27.55 &
23.53 &
30.13 &
31.87 &
27.87 \\
& U &
\textbf{27.71}$^{\ddag\ddag}$ &
\textbf{23.80}$^{\ddag}$ &
\textbf{30.54}$^{\ddag\ddag\dag\dag}$ &
\textbf{32.01}$^{\ddag}$ &
\textbf{28.15}$^{\ddag\ddag\dag\dag}$ \\
\bottomrule
\end{tabular}
\end{center}
\caption{\label{tab: main ende} 
BLEU scores on the WMT14 English-German translation task.
The ratio of authentic data to synthetic data is 1:1. 
\textproc{None}: only the authentic bilingual corpus is used.
\textproc{Search}: the translations of the monolingual corpus are generated by beam search \cite{Sennrich:16}.
\textproc{Sample}: the translations of the monolingual corpus are generated by sampling \cite{Edunov:18}.
``CE'': confidence estimation method.
``U'': uncertainty-based confidence estimation.
``N'': \textproc{NeuralQe}.
``All'': the combination of all test sets.
``$++$'': significantly better than \textproc{Search} without CE ($p<0.01$). 
``$*$'': significantly better than ``\textproc{Search} + N'' ($p < 0.05$).
``$\ddag$'': significantly better than \textproc{Sample} without CE ($p<0.05$).
``$\ddag\ddag$'': significantly better than \textproc{Sample} without CE ($p<0.01$).
``$\dag\dag$'': significantly better than ``\textproc{Sample} + N'' ($p < 0.01$).
}
\end{table*}


\subsection{Effect of Training Corpus Size}

\begin{figure}[!t]
    \centering
    \includegraphics[width=0.48\textwidth]{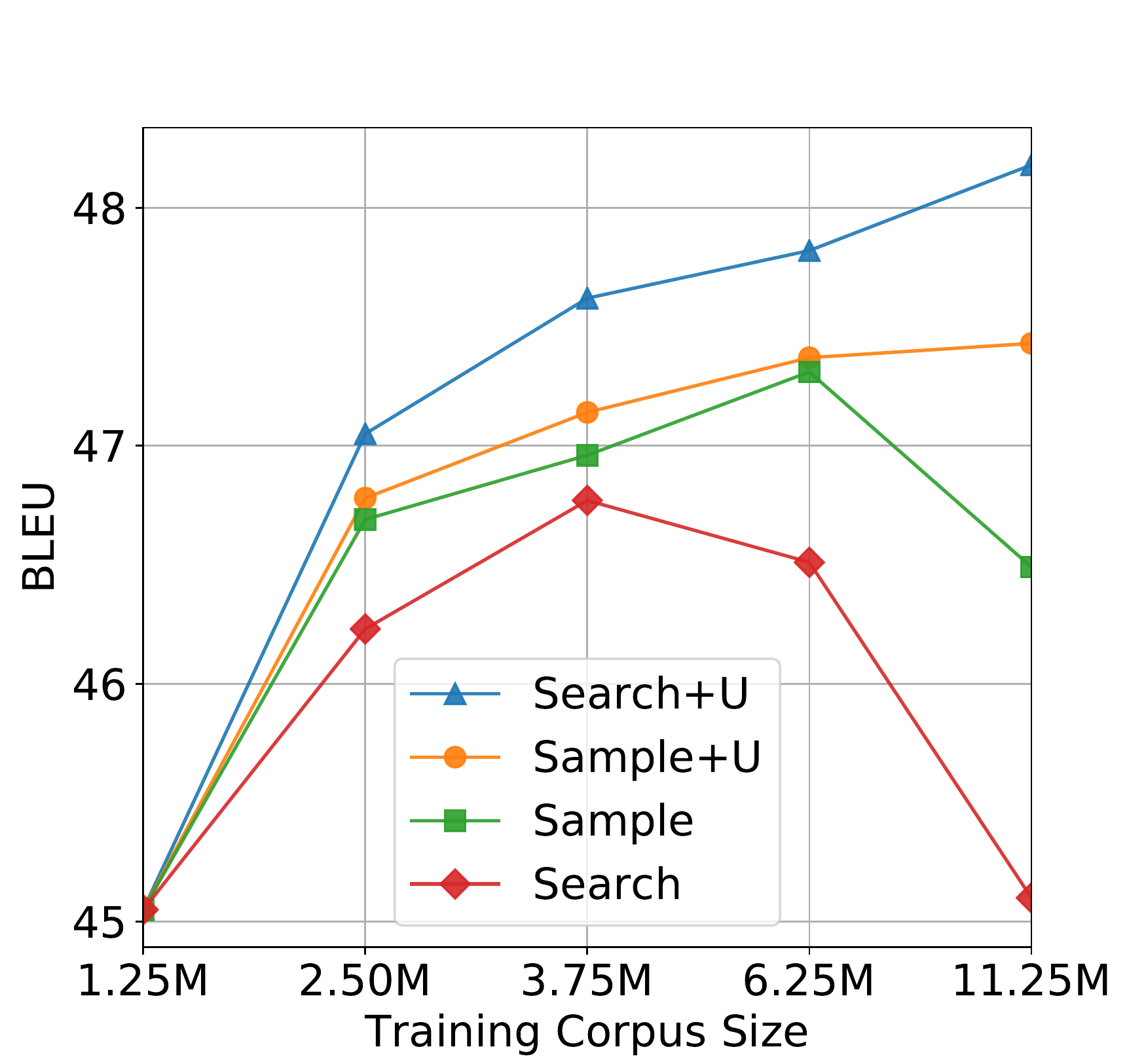}
    \caption{Effect of training corpus size.}
    \label{fig:corpus_size}
\end{figure}

Figure \ref{fig:corpus_size} shows the effect of training corpus size. The X-axis is the size of the total training data (i.e., $\mathcal{D}_b \cup \tilde{\mathcal{D}}_b$ in Eq. (\ref{eq:third_step})). The BLEU scores were calculated on the Chinese-English development set. We find that the translation performance of \textproc{Search} rises with the increase of monolingual corpus size in the beginning. However, further enlarging the monolingual corpus hurts the translation performance. In contrast, our approach  can still obtain further improvements when adding more synthetic bilingual sentence pairs. 
Similar findings are also observed for \textproc{Sample}.

\subsection{Effect of Data Selection}

Instead of randomly selecting monolingual sentences to generate synthetic data, we also used the method proposed by \cite{Fadaee:18} to select monolingual data by targeting difficult words. In this series of experiments, we used the same amount of monolingual data that was derived from a larger monolingual corpus using different data selection methods.

Results on NIST06 show that targeting difficult words improves over randomly selecting monolingual data (46.23 $\rightarrow$ 46.60 BLEU), confirming the finding of \citet{Fadaee:18}.
Using uncertainty-based confidence can further improve the translation performance (46.60 $\rightarrow$ 47.18 BLEU), indicating that our approach can be combined with advanced data selection methods.


\subsection{Case Study}

\begin{figure*}[!t]
    \centering
    \includegraphics[width=0.94\textwidth]{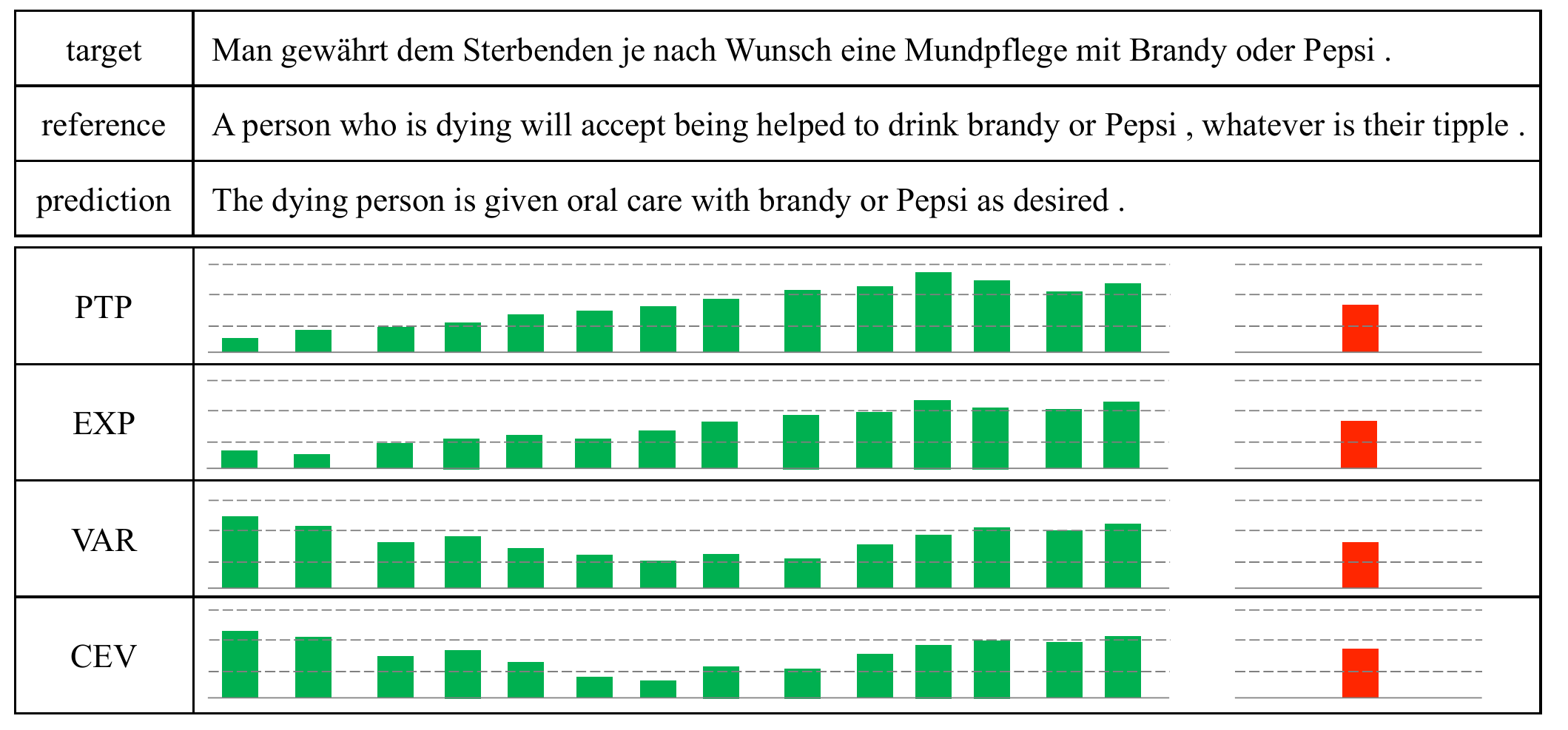}
    \caption{Example of confidence measures.}
    \label{fig:case}
\end{figure*}

Figure \ref{fig:case} shows an example of model prediction and its corresponding word- and sentence-level confidence measures for the English-German task. 
We observe that the PTP and EXP measures are unable to give low confidence to erroneous words. In contrast, variance-based measures such as VAR and CEV can better quantify how confident the model is to make its prediction.

\section{Related work}

Our work is closely related to three lines of research: (1) back-translation, (2) confidence estimation, and (3) uncertainty quantification.

\subsection{Back-translation}

Back-translation is a simple and effective approach to leveraging monolingual data for NMT \cite{Sennrich:16}. There has been a growing body of literature that analyzes and extends back-translation recently. \newcite{Currey:17} show that low-resource NMT can benefit from the synthetic data generated by simply copying target monolingual data to the source side. \newcite{Kenji:18} and \newcite{Edunov:18} demonstrate that it is more effective to generate source sentences via sampling rather than beam search. 
\newcite{Cotterell2018ExplainingAG} and \newcite{Hoang:18} find that iterative back-translation can further improve the performance of NMT. 
\newcite{Fadaee:18} show that words with high predicted loss during training benefit most. Our work differs from existing methods in that we propose to use confidence estimation to enable back-translation to better cope with noisy synthetic data, which can be easily combined with previous works.
Our experiments show that both neural and uncertainty-based confidence estimation methods benefit back-translation.

\subsection{Confidence Estimation}

Estimating the confidence or quality of the output of MT systems \cite{Ueffing:07,Specia2009ImprovingTC,Bach:11GoodnessAM,Salehi2014ConfidenceEF,Rikters:17,openkiwi} is important for enabling downstream applications such as post-editing and interactive MT to better cope with translation mistakes. While existing methods rely on external models to estimate confidence, our approach leverages model uncertainty to derive confidence measures. The major benefit is that our approach does not need labeled data.

\subsection{Uncertainty Quantification}

Reliable uncertainty quantification is key to building a robust artificial intelligent system. It has been successfully applied to many fields, including computer vision \cite{Kendall:15,Kendall:17}, time series prediction \cite{Zhu:17}, and natural language processing \cite{Dong:18,Xiao:19}. Our work differs from previous work in that we are interested in calculating uncertainty after the model has made the prediction rather during inference. \newcite{Ott2018AnalyzingUI} also analyze the inherent uncertainty of machine translation. The difference is that they focus on the existence of multiple correct translations for a single sentence while we aim to quantify the uncertainty of NMT models.

\section{Conclusions}

We have presented a method for qualifying model uncertainty for neural machine translation and use uncertainty-based confidence measures to improve back-translation. 
The key idea is to use Monte Carlo Dropout to sample translation probabilities to calculate model uncertainty, without the need for manually labeled data.
As our approach is transparent to model architectures, we plan to further verify the effectiveness of our approach on other downstream applications of NMT such as post-editing and interactive MT in the future.

\section*{Acknowledgments}
We thank all anonymous reviewers for their valuable comments. This work is supported by the National Key R\&D Program of China
(No. 2017YFB0202204), National Natural Science Foundation of China (No. 61761166008, No. 61432013), Beijing Advanced Innovation Center
for Language Resources (No. TYR17002), and the NExT++ project supported by the National Research Foundation, Prime Ministers Office, Singapore under its IRC@Singapore Funding Initiative. This research is also supported by Sogou Inc.

\bibliographystyle{acl_natbib}
\bibliography{emnlp-ijcnlp-2019}

\begin{thebibliography}{53}
\expandafter\ifx\csname natexlab\endcsname\relax\def\natexlab#1{#1}\fi

\bibitem[{Bach et~al.(2011)Bach, Huang, and Al-Onaizan}]{Bach:11GoodnessAM}
Nguyen Bach, Fei Huang, and Yaser Al-Onaizan. 2011.
\newblock \href {https://www.aclweb.org/anthology/P11-1022} {{G}oodness: A
  method for measuring machine translation confidence}.
\newblock In \emph{Proceedings of the 49th Annual Meeting of the Association
  for Computational Linguistics: Human Language Technologies}, pages 211--219,
  Portland, Oregon, USA. Association for Computational Linguistics.

\bibitem[{Bahdanau et~al.(2015)Bahdanau, Cho, and Bengio}]{Bahdanau:15}
Dzmitry Bahdanau, Kyunghyun Cho, and Yoshua Bengio. 2015.
\newblock Neural machine translation by jointly learning to align and
  translate.
\newblock In \emph{Proceedings of ICLR 2015}.

\bibitem[{Blatz et~al.(2004)Blatz, Fitzgerald, Foster, Gandrabur, Goutte,
  Kulesza, Sanchis, and Ueffing}]{Blatz:04}
John Blatz, Erin Fitzgerald, George Foster, Simona Gandrabur, Cyril Goutte,
  Alex Kulesza, Alberto Sanchis, and Nicola Ueffing. 2004.
\newblock \href {https://www.aclweb.org/anthology/C04-1046} {Confidence
  estimation for machine translation}.
\newblock In \emph{Coling 2004: Proceedings of the 20th international
  conference on computational linguistics}, pages 315--321.

\bibitem[{Blundell et~al.(2015)Blundell, Cornebise, Kavukcuoglu, and
  Wierstra}]{Blundell:15}
Charles Blundell, Julien Cornebise, Koray Kavukcuoglu, and Daan Wierstra. 2015.
\newblock Weight uncertainty in neural networks.
\newblock In \emph{Proceedings of ICML 2015}.

\bibitem[{Buntine and Weigend(1991)}]{Buntine:91}
Wray Buntine and Andreas~S. Weigend. 1991.
\newblock Bayesian back-propagation.
\newblock \emph{Complex Systems}.

\bibitem[{Chen et~al.(2017)Chen, Liu, Cheng, and Li}]{Chen:17}
Yun Chen, Yang Liu, Yong Cheng, and Victor~O.K. Li. 2017.
\newblock \href {https://doi.org/10.18653/v1/P17-1176} {A teacher-student
  framework for zero-resource neural machine translation}.
\newblock In \emph{Proceedings of the 55th Annual Meeting of the Association
  for Computational Linguistics (Volume 1: Long Papers)}, pages 1925--1935,
  Vancouver, Canada. Association for Computational Linguistics.

\bibitem[{Cheng et~al.(2016)Cheng, Xu, He, He, Wu, Sun, and Liu}]{Cheng:16}
Yong Cheng, Wei Xu, Zhongjun He, Wei He, Hua Wu, Maosong Sun, and Yang Liu.
  2016.
\newblock \href {https://www.aclweb.org/anthology/P16-1185} {Semi-supervised
  learning for neural machine translation}.
\newblock In \emph{Proceedings of ACL 2016}.

\bibitem[{Cotterell and Kreutzer(2018)}]{Cotterell2018ExplainingAG}
Ryan Cotterell and Julia Kreutzer. 2018.
\newblock Explaining and generalizing back-translation through wake-sleep.
\newblock \emph{CoRR}, abs/1806.04402.

\bibitem[{Currey et~al.(2017)Currey, Miceli~Barone, and Heafield}]{Currey:17}
Anna Currey, Antonio~Valerio Miceli~Barone, and Kenneth Heafield. 2017.
\newblock \href {https://www.aclweb.org/anthology/W17-4715} {Copied monolingual
  data improves low-resource neural machine translation}.
\newblock In \emph{Proceedings of the Second Conference on Machine
  Translation}, pages 148--156. Association for Computational Linguistics.

\bibitem[{Dong et~al.(2018)Dong, Quirk, and Lapata}]{Dong:18}
Li~Dong, Chris Quirk, and Mirella Lapata. 2018.
\newblock \href {https://doi.org/10.18653/v1/P18-1069} {Confidence modeling for
  neural semantic parsing}.
\newblock In \emph{Proceedings of the 56th Annual Meeting of the Association
  for Computational Linguistics (Volume 1: Long Papers)}, pages 743--753,
  Melbourne, Australia. Association for Computational Linguistics.

\bibitem[{Edunov et~al.(2018)Edunov, Ott, Auli, and Grangier}]{Edunov:18}
Sergey Edunov, Myle Ott, Michael Auli, and David Grangier. 2018.
\newblock \href {https://aclweb.org/anthology/D18-1045} {Understanding
  back-translation at scale}.
\newblock In \emph{Proceedings of EMNLP 2018}.

\bibitem[{Fadaee et~al.(2017)Fadaee, Bisazza, and Monz}]{Fadaee:17}
Marzieh Fadaee, Arianna Bisazza, and Christof Monz. 2017.
\newblock \href {https://doi.org/10.18653/v1/P17-2090} {Data augmentation for
  low-resource neural machine translation}.
\newblock In \emph{Proceedings of the 55th Annual Meeting of the Association
  for Computational Linguistics (Volume 2: Short Papers)}, pages 567--573,
  Vancouver, Canada. Association for Computational Linguistics.

\bibitem[{Fadaee and Monz(2018)}]{Fadaee:18}
Marzieh Fadaee and Christof Monz. 2018.
\newblock \href {https://www.aclweb.org/anthology/D18-1040} {Back-translation
  sampling by targeting difficult words in neural machine translation}.
\newblock In \emph{Proceedings of EMNLP 2018}.

\bibitem[{Gal and Ghahramani(2016)}]{Gal:16}
Yarin Gal and Zoubin Ghahramani. 2016.
\newblock Dropout as a bayesian approaximation: Representing model uncertainty
  in deep learning.
\newblock In \emph{Proceedings of ICML 2016}.

\bibitem[{Geifman et~al.(2019)Geifman, Uziel, and
  El-Yaniv}]{geifman:18biasreduced}
Yonatan Geifman, Guy Uziel, and Ran El-Yaniv. 2019.
\newblock \href {https://openreview.net/forum?id=SJfb5jCqKm} {Bias-reduced
  uncertainty estimation for deep neural classifiers}.
\newblock In \emph{International Conference on Learning Representations}.

\bibitem[{Graves(2011)}]{Graves:11PracticalVI}
Alex Graves. 2011.
\newblock Practical variational inference for neural networks.
\newblock In \emph{Proceedings of NeurIPS}.

\bibitem[{Hassan et~al.(2018)Hassan, Aue, Chen, Chowdhary, Clark, Federmann,
  Huang, Junczys-Dowmunt, Lewis, Li, Liu, Liu, Luo, Menezes, Qin, Seide, Tan,
  Tian, Wu, Wu, Xia, Zhang, Zhang, and Zhou}]{Hassan:18}
Hany Hassan, Anthony Aue, Chang Chen, Vishal Chowdhary, Jonathan Clark,
  Christian Federmann, Xuedong Huang, Marcin Junczys-Dowmunt, William Lewis,
  Mu~Li, Shujie Liu, Tie-Yan Liu, Renqian Luo, Arul Menezes, Tao Qin, Frank
  Seide, Xu~Tan, Fei Tian, Lijun Wu, Shuangzhi Wu, Yingce Xia, Dongdong Zhang,
  Zhirui Zhang, and Ming Zhou. 2018.
\newblock Achieving human parity on automatic chinese to english news
  translation.
\newblock arXiv:1803.05567.

\bibitem[{Hoang et~al.(2018)Hoang, Keohn, Haffari, and Cohn}]{Hoang:18}
Cong Duy~Vu Hoang, Philipp Keohn, Gholamreza Haffari, and Trevor Cohn. 2018.
\newblock \href {https://www.aclweb.org/anthology/W18-2703} {Iterative
  back-translation for neural machine translation}.
\newblock In \emph{Proceedings of the 2nd Workshop on Neural Machine
  Translation and Generation}.

\bibitem[{Imamura et~al.(2018)Imamura, Fujita, and Sumita}]{Kenji:18}
Kenji Imamura, Atsushi Fujita, and Eiichiro Sumita. 2018.
\newblock \href {https://www.aclweb.org/anthology/W18-2707} {Enhancement of
  encoder and attention using target monolingual corpora in neural machine
  translation}.
\newblock In \emph{Proceedings of the 2nd Workshop on Neural Machine
  Translation and Generation}, pages 55--63. Association for Computational
  Linguistics.

\bibitem[{Ive et~al.(2018)Ive, Blain, and Specia}]{Ive:18}
Julia Ive, Fr{\'e}d{\'e}ric Blain, and Lucia Specia. 2018.
\newblock \href {https://www.aclweb.org/anthology/C18-1266} {deep{Q}uest: A
  framework for neural-based quality estimation}.
\newblock In \emph{Proceedings of the 27th International Conference on
  Computational Linguistics}, pages 3146--3157, Santa Fe, New Mexico, USA.
  Association for Computational Linguistics.

\bibitem[{Kendall et~al.(2015)Kendall, Badrinarayanan, and
  Cipolla}]{Kendall:15}
Alex Kendall, Vijay Badrinarayanan, and Roberto Cipolla. 2015.
\newblock Bayesian segnet: Model uncertainty in deep convolutional
  encoder-decoder architectures for scene understanding.
\newblock arXiv:1511.02680.

\bibitem[{Kendall and Gal(2017)}]{Kendall:17}
Alex Kendall and Yarin Gal. 2017.
\newblock What uncertainties do we need in bayesian deep learning for computer
  vision?
\newblock In \emph{Advances in neural information processing systems}, pages
  5574--5584.

\bibitem[{Kepler et~al.(2019)Kepler, Tr{\'e}nous, Treviso, Vera, and
  Martins}]{openkiwi}
Fabio Kepler, Jonay Tr{\'e}nous, Marcos Treviso, Miguel Vera, and Andr{\'e}
  F.~T. Martins. 2019.
\newblock \href {https://www.aclweb.org/anthology/P19-3020} {{O}pen{K}iwi: An
  open source framework for quality estimation}.
\newblock In \emph{Proceedings of the 57th Annual Meeting of the Association
  for Computational Linguistics: System Demonstrations}, pages 117--122,
  Florence, Italy. Association for Computational Linguistics.

\bibitem[{Kim et~al.(2017)Kim, Lee, and Na}]{Kim:17}
Hyun Kim, Jong-Hyeok Lee, and Seung-Hoon Na. 2017.
\newblock \href {https://doi.org/10.18653/v1/W17-4763} {Predictor-estimator
  using multilevel task learning with stack propagation for neural quality
  estimation}.
\newblock In \emph{Proceedings of the Second Conference on Machine
  Translation}, pages 562--568, Copenhagen, Denmark. Association for
  Computational Linguistics.

\bibitem[{Kingma and Ba(2015)}]{Kingma2015AdamAM}
Diederik~P. Kingma and Jimmy Ba. 2015.
\newblock Adam: A method for stochastic optimization.
\newblock \emph{CoRR}, abs/1412.6980.

\bibitem[{Koehn(2004)}]{Koehn2004Statistical}
Philipp Koehn. 2004.
\newblock \href {https://www.aclweb.org/anthology/W04-3250} {Statistical
  significance tests for machine translation evaluation}.
\newblock In \emph{Proceedings of the 2004 conference on empirical methods in
  natural language processing}, pages 388--395.

\bibitem[{Koehn et~al.(2007)Koehn, Hoang, Birch, Callison-Burch, Federico,
  Bertoldi, Cowan, Shen, Moran, Zens, Dyer, Bojar, Constantin, and
  Herbst}]{Koehn2007MosesOS}
Philipp Koehn, Hieu Hoang, Alexandra Birch, Chris Callison-Burch, Marcello
  Federico, Nicola Bertoldi, Brooke Cowan, Wade Shen, Christine Moran, Richard
  Zens, Chris Dyer, Ondrej Bojar, Alexandra Constantin, and Evan Herbst. 2007.
\newblock \href {https://www.aclweb.org/anthology/P07-2045} {Moses: Open source
  toolkit for statistical machine translation}.
\newblock In \emph{Proceedings of ACL 2007}.

\bibitem[{Koehn et~al.(2003)Koehn, Och, and Marcu}]{Koehn:03}
Philipp Koehn, Franz~J. Och, and Daniel Marcu. 2003.
\newblock \href {https://www.aclweb.org/anthology/N03-1017} {Statistical
  phrase-based translation}.
\newblock In \emph{Proceedings of NAACL 2003}.

\bibitem[{Lample et~al.(2018)Lample, Ott, Conneau, Denoyer, and
  Ranzato}]{Lample:18}
Guillaume Lample, Myle Ott, Alexis Conneau, Ludovic Denoyer, and Marc'Aurelio
  Ranzato. 2018.
\newblock \href {https://aclweb.org/anthology/D18-1549} {Phrase-based \& neural
  unsupervised machine translation}.
\newblock In \emph{Proceedings of EMNLP 2018}.

\bibitem[{Lee et~al.(2019)Lee, Hou, Mandalika, Lee, and
  Srinivasa}]{lee:18bayesian}
Gilwoo Lee, Brian Hou, Aditya Mandalika, Jeongseok Lee, and Siddhartha~S.
  Srinivasa. 2019.
\newblock \href {https://openreview.net/forum?id=SJGvns0qK7} {Bayesian policy
  optimization for model uncertainty}.
\newblock In \emph{International Conference on Learning Representations}.

\bibitem[{Luong et~al.(2017)Luong, Besacier, and Lecouteux}]{Luong2017FindTE}
Ngoc-Quang Luong, Laurent Besacier, and Benjamin Lecouteux. 2017.
\newblock Find the errors, get the better: Enhancing machine translation via
  word confidence estimation.
\newblock \emph{Natural Language Engineering}, 23:617--639.

\bibitem[{Oh et~al.(2019)Oh, Gallagher, Murphy, Schroff, Pan, and
  Roth}]{oh:18modeling}
Seong~Joon Oh, Andrew~C. Gallagher, Kevin~P. Murphy, Florian Schroff, Jiyan
  Pan, and Joseph Roth. 2019.
\newblock \href {https://openreview.net/forum?id=r1xQQhAqKX} {Modeling
  uncertainty with hedged instance embeddings}.
\newblock In \emph{International Conference on Learning Representations}.

\bibitem[{Ott et~al.(2018)Ott, Auli, Grangier, and
  Ranzato}]{Ott2018AnalyzingUI}
Myle Ott, Michael Auli, David Grangier, and Marc'Aurelio Ranzato. 2018.
\newblock Analyzing uncertainty in neural machine translation.
\newblock In \emph{Proceedings of ICML 2018}.

\bibitem[{Papineni et~al.(2001)Papineni, Roukos, Ward, and
  Zhu}]{Papineni2001BleuAM}
Kishore Papineni, Salim Roukos, Todd Ward, and Wei-Jing Zhu. 2001.
\newblock \href {https://www.aclweb.org/anthology/P02-1040} {Bleu: a method for
  automatic evaluation of machine translation}.
\newblock In \emph{Proceedings of ACL 2001}.

\bibitem[{Pereyra et~al.(2017)Pereyra, Tucker, Chorowski, Kaiser, and
  Hinton}]{Pereyra:2017}
Gabriel Pereyra, George Tucker, Jan Chorowski, Lukasz Kaiser, and Geoffrey~E.
  Hinton. 2017.
\newblock Regularizing neural networks by penalizing confident output
  distributions.
\newblock \emph{CoRR}, abs/1701.06548.

\bibitem[{Pinnis et~al.(2017)Pinnis, Kri{\v{s}}lauks, Deksne, and
  Miks}]{Pinnis:17}
M{\={a}}rcis Pinnis, Rihards Kri{\v{s}}lauks, Daiga Deksne, and Toms Miks.
  2017.
\newblock Neural machine translation for morphologically rich languages with
  improved sub-word units and synthetic data.
\newblock In \emph{Text, Speech, and Dialogue}, pages 237--245, Cham. Springer
  International Publishing.

\bibitem[{Poncelas et~al.(2018)Poncelas, Shterionov, Way, de~Buy~Wenniger, and
  Passban}]{Poncelas:2018}
Alberto Poncelas, Dimitar Shterionov, Andy Way, Gideon~Maillette
  de~Buy~Wenniger, and Peyman Passban. 2018.
\newblock Investigating backtranslation in neural machine translation.
\newblock \emph{CoRR}, abs/1804.06189.

\bibitem[{Ren et~al.(2018)Ren, Chen, Liu, Li, Zhou, and Ma}]{Ren:18}
Shuo Ren, Wenhu Chen, Shujie Liu, Mu~Li, Ming Zhou, and Shuai Ma. 2018.
\newblock \href {https://doi.org/10.18653/v1/P18-1006} {Triangular architecture
  for rare language translation}.
\newblock In \emph{Proceedings of the 56th Annual Meeting of the Association
  for Computational Linguistics (Volume 1: Long Papers)}, pages 56--65,
  Melbourne, Australia. Association for Computational Linguistics.

\bibitem[{Rikters and Fishel(2017)}]{Rikters:17}
Matiss Rikters and Mark Fishel. 2017.
\newblock Confidence through attention.
\newblock \emph{CoRR}, abs/1710.03743.

\bibitem[{Salehi et~al.(2014)Salehi, Khadivi, and
  Riahi}]{Salehi2014ConfidenceEF}
Marzieh Salehi, Shahram Khadivi, and Nooshin Riahi. 2014.
\newblock Confidence estimation for machine translation using context vectors.
\newblock \emph{7'th International Symposium on Telecommunications (IST'2014)},
  pages 524--528.

\bibitem[{Sennrich et~al.(2016{\natexlab{a}})Sennrich, Haddow, and
  Birch}]{Sennrich:16}
Rico Sennrich, Barry Haddow, and Alexandra Birch. 2016{\natexlab{a}}.
\newblock \href {https://www.aclweb.org/anthology/P16-1009} {Improving neural
  machine translation models with monolingual data}.
\newblock In \emph{Proceedings of ACL 2016}.

\bibitem[{Sennrich et~al.(2016{\natexlab{b}})Sennrich, Haddow, and
  Birch}]{Sennrich2016NeuralMT}
Rico Sennrich, Barry Haddow, and Alexandra Birch. 2016{\natexlab{b}}.
\newblock \href {https://www.aclweb.org/anthology/P16-1162} {Neural machine
  translation of rare words with subword units}.
\newblock In \emph{Proceedings of the 54th Annual Meeting of the Association
  for Computational Linguistics (Volume 1: Long Papers)}, pages 1715--1725.

\bibitem[{Specia et~al.(2009)Specia, Turchi, Wang, Shawe-Taylor, and
  Saunders}]{Specia2009ImprovingTC}
Lucia Specia, Marco Turchi, Zhuoran Wang, John Shawe-Taylor, and Craig
  Saunders. 2009.
\newblock Improving the confidence of machine translation quality estimates.
\newblock In \emph{Twelfth Machine Translation Summit}.

\bibitem[{Sutskever et~al.(2014)Sutskever, Vinyals, and Le}]{Sutskever:14}
Ilya Sutskever, Oriol Vinyals, and Quoc~V. Le. 2014.
\newblock Sequence to sequence learning with neural networks.
\newblock In \emph{Proceedings of NeurIPS 2014}.

\bibitem[{Szegedy et~al.(2016)Szegedy, Vanhoucke, Ioffe, Shlens, and
  Wojna}]{Szegedy:2016}
Christian Szegedy, Vincent Vanhoucke, Sergey Ioffe, Jonathon Shlens, and
  Zbigniew Wojna. 2016.
\newblock Rethinking the inception architecture for computer vision.
\newblock \emph{2016 IEEE Conference on Computer Vision and Pattern Recognition
  (CVPR)}, pages 2818--2826.

\bibitem[{Ueffing and Ney(2007)}]{Ueffing:07}
Nicola Ueffing and Hermann Ney. 2007.
\newblock \href {https://www.aclweb.org/anthology/J07-1003} {Word-level
  confidence estimation for machine translation}.
\newblock \emph{Computational Linguistics}.

\bibitem[{Vaswani et~al.(2017)Vaswani, Shazeer, Parmar, Uszkoreit, Jones,
  Gomez, Kaiser, and Polosukhin}]{Vaswani:17}
Ashish Vaswani, Noam Shazeer, Niki Parmar, Jakob Uszkoreit, Llion Jones,
  Aidan~N. Gomez, Lukasz Kaiser, and Illia Polosukhin. 2017.
\newblock Attention is all you need.
\newblock In \emph{Proceedings of NeurIPS 2017}.

\bibitem[{Wang et~al.(2018)Wang, Fan, Li, Zhou, Chen, Shi, and Si}]{Wang:18}
Jiayi Wang, Kai Fan, Bo~Li, Fengming Zhou, Boxing Chen, Yangbin Shi, and Luo
  Si. 2018.
\newblock \href {https://doi.org/10.18653/v1/W18-6465} {{A}libaba submission
  for {WMT}18 quality estimation task}.
\newblock In \emph{Proceedings of the Third Conference on Machine Translation:
  Shared Task Papers}, pages 809--815, Belgium, Brussels. Association for
  Computational Linguistics.

\bibitem[{Wu et~al.(2016)Wu, Schuster, Chen, Le, Norouzi, Macherey, Krikun,
  Cao, Gao, Macherey, Klingner, Shah, Johnson, Liu, Kaiser, Gouws, Kato, Kudo,
  Kazawa, Stevens, Kurian, Patil, Wang, Young, Smith, Riesa, Rudnick, Vinyals,
  Corrado, Hughes, and Dean}]{Wu:16}
Yonghui Wu, Mike Schuster, Zhifeng Chen, Quoc~V. Le, Mohammad Norouzi, Wolfgang
  Macherey, Maxim Krikun, Yuan Cao, Qin Gao, Klaus Macherey, Jeff Klingner,
  Apurva Shah, Melvin Johnson, Xiaobing Liu, Lukasz Kaiser, Stephan Gouws,
  Yoshikiyo Kato, Taku Kudo, Hideto Kazawa, Keith Stevens, George Kurian,
  Nishant Patil, Wei Wang, Cliff Young, Jason Smith, Jason Riesa, Alex Rudnick,
  Oriol Vinyals, Greg Corrado, Macduff Hughes, and Jeffrey Dean. 2016.
\newblock Google's neural machine translation system: Bridging the gap between
  human and machine translation.
\newblock arXiv:1609.08144v2.

\bibitem[{Xiao and Wang(2019)}]{Xiao:19}
Yijun Xiao and William~Yang Wang. 2019.
\newblock Quantifying uncertainties in natural language processing tasks.
\newblock In \emph{Proceedings of AAAI 2019}.

\bibitem[{Zhang et~al.(2017)Zhang, Ding, Shen, Cheng, Sun, Luan, and
  Liu}]{THUMT}
Jiacheng Zhang, Yanzhuo Ding, Shiqi Shen, Yong Cheng, Maosong Sun, Huan{-}Bo
  Luan, and Yang Liu. 2017.
\newblock \href {http://arxiv.org/abs/1706.06415} {{THUMT:} an open source
  toolkit for neural machine translation}.
\newblock \emph{CoRR}, abs/1706.06415.

\bibitem[{Zhu and Laptev(2017)}]{Zhu:17}
Lingxue Zhu and Nikolay Laptev. 2017.
\newblock Deep and confident prediction for time series at uber.
\newblock In \emph{Proceedings of ICDM 2017}.

\bibitem[{Zoph et~al.(2016)Zoph, Yuret, May, and Knight}]{Zoph:16}
Barret Zoph, Deniz Yuret, Jonathan May, and Kevin Knight. 2016.
\newblock \href {https://aclweb.org/anthology/D16-1163} {Transfer learning for
  low-resource neural machine translation}.
\newblock In \emph{Proceedings of EMNLP 2016}.

\end{thebibliography}

\end{document}